\documentclass{article}
\usepackage{amssymb}
\usepackage{pifont}
\usepackage{cite}
\usepackage{bookmark}
\usepackage{amsmath,amssymb,amsfonts}
\usepackage{algorithmic}
\usepackage{graphicx}
\usepackage{textcomp}
\usepackage{url}
\usepackage{hyperref}
\usepackage{float}
\usepackage{siunitx}
\usepackage{spconf,amsmath,graphicx,hyperref}
\usepackage{enumitem}
\usepackage{booktabs,threeparttable,makecell}
\usepackage{xcolor,multirow}
\usepackage{bm}
\usepackage[dvipsnames]{xcolor}   
\usepackage{pifont}               
\definecolor{ForestGreen}{RGB}{34,139,34}
\definecolor{BrickRed}{RGB}{203,65,84}

\usepackage{caption}        
\title{Structure-Aware Contrastive Learning with Fine-Grained Binding Representations for Drug Discovery}
\name{
    \begin{tabular}{c}
      Jing Lan\textsuperscript{*1}\thanks{*Co-first author}, 
      Hexiao Ding\textsuperscript{*1}, 
      Hongzhao Chen\textsuperscript{*1}, 
      Yufeng Jiang\textsuperscript{1}, 
      Nga-Chun Ng\textsuperscript{1,2}, \\ 
      Gwing Kei Yip\textsuperscript{1,3},
      Gerald W.Y. Cheng\textsuperscript{1}, 
      Yunlin Mao\textsuperscript{1}, 
      Jing Cai\textsuperscript{1}, 
      Liang-ting Lin\textsuperscript{1}, 
      Jung Sun Yoo\textsuperscript{\#1}\thanks{\#Corresponding author}
    \end{tabular}
}
\address{\textsuperscript{1}Department of Health Technology and Informatics, The Hong Kong Polytechnic University\\
    \textsuperscript{2}Department of Nuclear Medicine and PET, Hong Kong Sanatorium and Hospital\\
    \textsuperscript{3}Nuclear Medicine Unit, Department of Diagnostic and Interventional Radiology, Queen Elizabeth Hospital \\
    Hong Kong SAR, China\\
    Emails: \{jing-hti.lan, hexiao.ding, hongzhao.chen, yufeng.jiang, yunlin.mao\}@connect.polyu.hk\\
    \{wai-yeung.cheng, jing.cai, ltlin, jungsun.yoo\}@polyu.edu.hk\\
    \{sam.nc.ng\}@hksh.com, \{gwinky.yip\}@ha.org.hk}
\begin{document}
\maketitle
\begin{abstract}

Accurate identification of drug–target interactions (DTI) remains a central challenge in computational pharmacology, where sequence-based methods offer scalability. This work introduces a sequence-based drug–target interaction framework that integrates structural priors into protein representations while maintaining high-throughput screening capability. Evaluated across multiple benchmarks, the model achieves state-of-the-art performance on Human and BioSNAP datasets and remains competitive on BindingDB. In virtual screening tasks, it surpasses prior methods on LIT-PCBA, yielding substantial gains in AUROC and BEDROC. Ablation studies confirm the critical role of learned aggregation, bilinear attention, and contrastive alignment in enhancing predictive robustness. Embedding visualizations reveal improved spatial correspondence with known binding pockets and highlight interpretable attention patterns over ligand–residue contacts. These results validate the framework’s utility for scalable and structure-aware DTI prediction. Code is available at \url{https://github.com/1anj/SaBAN-DTI}.
\end{abstract}

\section{Introduction}
\label{sec:intro}

Predicting compound–protein binding potential is essential for drug discovery. However, binding affinity alone does not capture the full pharmacological features because location of binding, conformational adaptation, and kinetic stability shape efficacy and selectivity \cite{Copeland2016}. Physics-based free energy methods can fully recover these factors yet remain costly at the screening scale \cite{York2023}. Fast geometric docking has improved throughput, but the practical gate in many pipelines is still the presence or absence of interaction. 
Therefore, drug-target interaction (DTI) prediction functions as a rapid prescreen that precedes pose refinement and affinity estimation.

Public datasets and benchmarks have advanced DTI research by enabling model development and exposing persistent gaps. ChEMBL \cite{Zdrazil2025} provides extensive bioactivity annotations across diverse targets and modalities, BioLiP2 \cite{Zhang2024} integrates structure-based evidence for biologically relevant interactions. Moreover, BindingDB \cite{Liu2025} continues to expand with curated binding measurements from literature and patents. Despite these resources, a considerable portion of interaction evidence remains embedded in unstructured text and requires specialized extraction methods such as DrugProt \cite{Miranda2023} to produce machine-readable data. These approaches enable efficient training and provide interpretable attention-based explanations. However, they remain sensitive to dataset split strategies and often make limited use of explicit pocket level geometric information.

We present a sequence-based DTI framework that injects structural signals while preserving screening speed and accuracy. Proteins are represented using a structure-aware vocabulary that pairs each residue token with a compact descriptor of local geometry. 
A large protein language model is pretrained on sequences annotated with these descriptors, enabling the encoder to learn structural context while operating on plain sequence inputs. 
Encoding small molecules with SELFIES guarantees validity and preserves chemical semantics \cite{Krenn2022}. 
Both drug and protein encoders employ a patch-based convolutional trunk that maintains input resolution and supplies token features to attention-based aggregation, which yields an interpretable importance map over patches \cite{su2023saprot}. A contrastive learning objective aligns drug and protein embeddings by drawing together regions corresponding to true binding interfaces. Our main contributions are as follows:
\begin{enumerate}[nosep]
  \item Propose a structure-aware protein sequence representation that augments each residue token with a compact local-geometry descriptor, enabling structural context learning from plain sequences.
  \item Introduce an attention-based aggregation module that preserves resolution, focuses on binding-relevant regions, and produces interpretable importance maps.
  \item Deliver a DTI prediction framework that outperforms existing baselines in accuracy and speed, supporting large-scale virtual screening for drug discovery.
\end{enumerate}
\section{Methodology}
\label{sec:method}

\begin{figure}[htbp]
    \includegraphics[width=\linewidth]{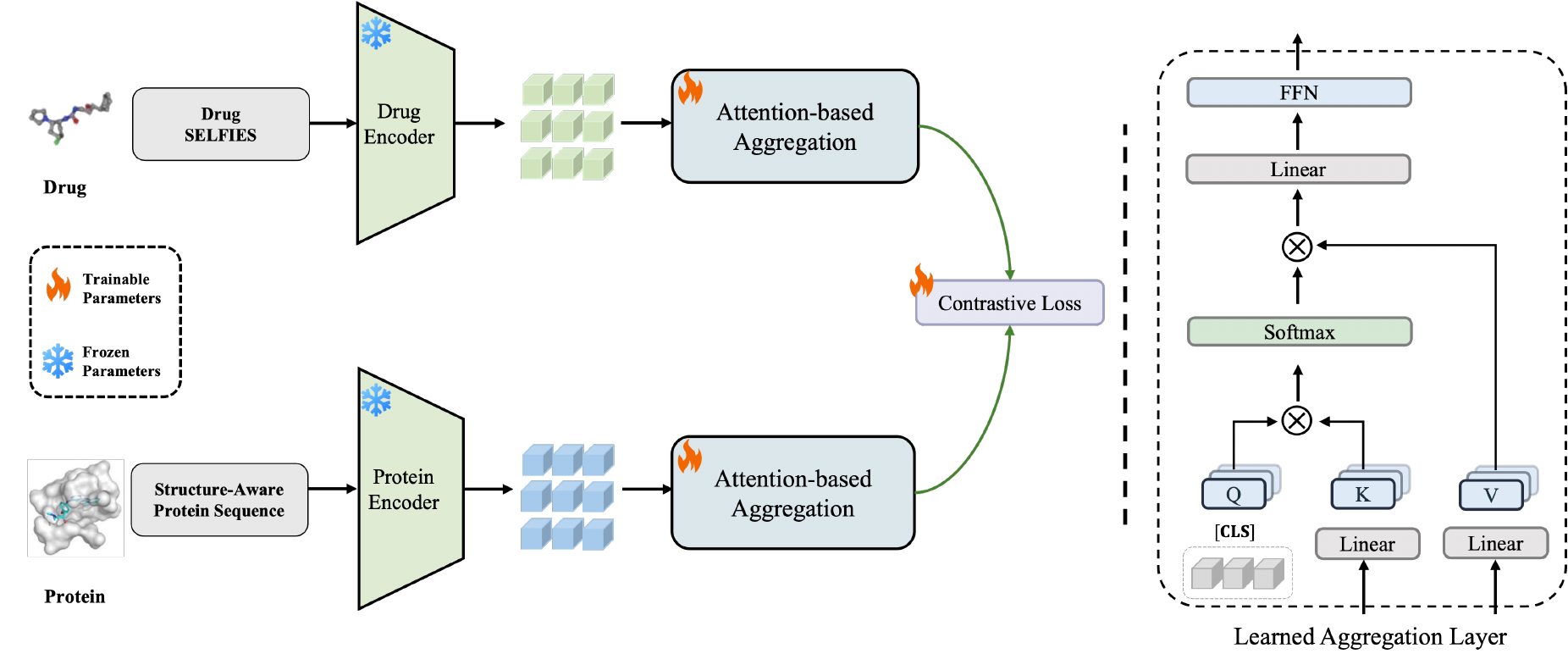}
    \caption{\textbf{Model Architecture.} Our framework encodes drug and protein sequences independently using frozen pre-trained models. The resulting embeddings are aggregated through an attention-based module and optimized via a contrastive learning objective. Cosine similarity between drug and protein representations is used to compute interaction probabilities, which support discriminative prediction of drug-target interactions and enable large-scale virtual screening.
    }
    \label{fig:framework}
\end{figure}

\subsection{Representations of Drugs and Protein Targets}
\subsubsection{Drug and protein encoders} 

As shown in Fig.~\ref{fig:framework}, the initial inputs of our model are string-based representations of drugs and proteins. For a protein sequence $P$, we adopt the \emph{structure-aware vocabulary}~\cite{su2023saprot}, where each residue is replaced by one of 441 tokens that couple amino acids with 3-D geometric features. This design alleviates the lack of explicit structural information in amino acid sequences. For a drug sequence $D$, we employ SELFIES~\cite{Krenn2022}, a robust chemical string representation where every valid string representation can be unambiguously decoded into a chemically valid molecular graph, thereby avoiding invalid structures that may arise from SMILES.

Two frozen pre-trained encoders are employed to transform input sequences into representations. Protein sequences are encoded by SaProt~\cite{su2023saprot}, which jointly models amino acid tokens and structural features to produce sequence embeddings $\mathbf{H}_t$. Drug SELFIES sequences are processed by SELFormer~\cite{selformer}, yielding sequence embeddings $\mathbf{H}_d$. The encoders remain frozen during training to leverage their learned structural priors efficiently, while only the downstream interaction modules and projection layers are optimized. Our selection of encoders is grounded in controlled ablation studies.

\subsubsection{Attention-based pooling}  
Local binding pockets contain key drug-target interaction signals, which can be diluted when subjected to average pooling through a learned aggregation layer. To preserve fine-grained contact features encoded via self-attention, token-level embeddings are aggregated using an attention-based pooling module adapted from transformer architectures~\cite{Touvron2021}.

Formally, let $\mathbf{H}\in\mathbb{R}^{n\times d_h}$ denote the sequence of $n$ token embeddings (i.e., $\mathbf{H}_d$ or $\mathbf{H}_t$) with hidden size $d_h$. A learnable class token $\mathbf{q}\in\mathbb{R}^{1\times d_h}$ serves as the query against projected keys and values:
\begin{equation}
\mathbf{K} = \mathbf{H}\mathbf{W}_K, \quad
\mathbf{V} = \mathbf{H}\mathbf{W}_V, \qquad
\mathbf{W}_K, \mathbf{W}_V \in \mathbb{R}^{d_h \times d_h}.
\end{equation}

The attention weights and pooled embedding are computed as
\begin{equation}
\boldsymbol{\alpha} = \mathrm{softmax}\!\left(\tfrac{\mathbf{q}\mathbf{K}^{\top}}{\sqrt{d_h}}\right) \in \mathbb{R}^{1 \times n}, \qquad
\tilde{\mathbf{z}} = \boldsymbol{\alpha}\mathbf{V} \in \mathbb{R}^{1 \times d_h}.
\end{equation}

A residual connection with the class token, followed by layer normalization and a lightweight feed-forward network (FFN) with dropout rate 0.05, produces the final representation:
\begin{equation}
\mathbf{h} = \mathrm{FFN}(\mathrm{LayerNorm}(\tilde{\mathbf{z}} + \mathbf{q})) \in \mathbb{R}^{1 \times d_h}.
\end{equation}

This formulation concentrates the aggregation into a single-head attention block, ensuring interpretability of how token-level signals contribute to the final representation.

\subsection{Contrastive Learning for Drug-Target Interactions}
We align drug and protein embeddings in a shared latent space via a symmetric contrastive objective. Given $N$ drug--target pairs, pooled embeddings $\mathbf{h}_d^i$ and $\mathbf{h}_t^i$ are projected to co-embeddings via learnable linear projections $C_d\colon\mathbb{R}^{d_h}\to\mathbb{R}^{d_{\mathrm{proj}}}$ and $C_t\colon\mathbb{R}^{d_h}\to\mathbb{R}^{d_{\mathrm{proj}}}$:
\begin{equation}
\mathbf{Z}_d^i = C_d(\mathbf{h}_d^i), \qquad \mathbf{Z}_t^i = C_t(\mathbf{h}_t^i),
\end{equation}
and normalized as $\hat{\mathbf{Z}}_d^i = \mathbf{Z}_d^i / \|\mathbf{Z}_d^i\|$, $\hat{\mathbf{Z}}_t^i = \mathbf{Z}_t^i / \|\mathbf{Z}_t^i\|$. Cosine similarity is thus given by the inner product $\hat{\mathbf{Z}}_d^i \cdot \hat{\mathbf{Z}}_t^j$.

For drug-to-target alignment, the probability of a correct match is
\begin{equation}
p(t=i \mid d=i) =
\frac{\exp\!\left(\tau\, \hat{\mathbf{Z}}_d^i \cdot \hat{\mathbf{Z}}_t^i\right)}
{\sum_{j=1}^N \exp\!\left(\tau\, \hat{\mathbf{Z}}_d^i \cdot \hat{\mathbf{Z}}_t^j\right)},
\end{equation}
where $\tau$ is a temperature scaling parameter (set to 1 in our experiments). An analogous probability $p(d=i \mid t=i)$ is computed for target-to-drug alignment. The overall loss is the symmetric InfoNCE:
\begin{equation}
\mathcal{L}_{\mathrm{con}} =
-\tfrac{1}{2N}\sum_{i=1}^N \big[
\log p(t=i \mid d=i) + \log p(d=i \mid t=i)
\big].
\end{equation}

This training objective pulls embeddings of interacting pairs together and pushes non-interacting pairs apart, yielding discriminative and transferable representations for DTI prediction and virtual screening.

\subsection{Bilinear Attention for Interaction Representation}
A bilinear attention network (BAN) captures token-level interactions between drug and target.
Given token embeddings $\mathbf{H}_d\!\in\!\mathbb{R}^{n_d\times d_h}$ and $\mathbf{H}_t\!\in\!\mathbb{R}^{n_t\times d_h}$,
we first obtain projected features $\mathbf{D}_b=\phi(\mathbf{H}_d\mathbf{W}_U)$ and $\mathbf{T}_b=\phi(\mathbf{H}_t\mathbf{W}_V)$
with learnable weights $\mathbf{W}_U,\mathbf{W}_V\!\in\!\mathbb{R}^{d_h\times r}$ and pointwise nonlinearity $\phi$.

\begin{equation}\label{eq:ban-attn}
\mathbf{A}=\mathrm{softmax}\!\big(\mathbf{D}_b\mathbf{T}_b^\top\big)\in\mathbb{R}^{n_d\times n_t}
\end{equation}
\begin{equation}\label{eq:ban-fuse}
\mathbf{f}=\sum_{i=1}^{n_d}\Big((\mathbf{D}_b)_i \odot (\mathbf{A}\mathbf{T}_b)_i\Big)\in\mathbb{R}^{r}
\end{equation}
where $\odot$ denotes element-wise multiplication.

A single-head BAN uses \eqref{eq:ban-attn}--\eqref{eq:ban-fuse}, and multi-glimpse BAN repeats them with separate
$(\mathbf{W}_{U,g},\mathbf{W}_{V,g})$ and concatenates $\mathbf{f}_g$.
Finally, an MLP maps $\mathbf{f}$ to the interaction probability.
\section{Experiments}
\label{sec:exp}

\subsection{Datasets and Baselines}
The framework is assessed using DTI prediction and virtual screening. Following the DrugBAN~\cite{bai2023interpretable} setting, BindingDB~\cite{Liu2025}, BioSNAP~\cite{zitnik2018biosnap}, and Human~\cite{liu2015credible,chen2020transformercpi} datasets are employed for DTI prediction, where each dataset is split into training, validation, and test sets with a 7:1:2 ratio, and evaluated using 5-fold cross-validation. Virtual screening is conducted on the LIT-PCBA benchmark~\cite{tran2020litpcba}. Dataset statistics are reported in Table~\ref{tab:datasets}.  Strict data separation is maintained to prevent leakage. For DTI tasks, we ensure training, validation, and test sets are disjoint at the pair level. For virtual screening, we follow the official split protocols based on target identity, ensuring unseen targets during evaluation.

The framework is evaluated against classical machine learning (ML) and deep learning (DL) baselines. Classical ML models include SVM and random forest trained on ECFP4~\cite{rogers2010extended} and PSC descriptors~\cite{cao2013propy}. DL baselines cover GNN-CPI~\cite{tsubaki2019compound}, DeepConv-DTI~\cite{lee2019deepconv}, GraphDTA~\cite{nguyen2021graphdta}, MolTrans~\cite{huang2021moltrans}, and DrugBAN~\cite{bai2023interpretable}. For virtual screening, comparisons include Surflex~\cite{spitzer2012surflex}, Glide-SP~\cite{halgren2004glide}, Planet~\cite{zhang2023planet}, BigBind~\cite{brocidiacono2022bigbind}, and DrugCLIP~\cite{drugclip}. Also, DrugCLIP is included as a contrastive learning baseline.

\subsection{Implementation Details}
The model is optimized via AdamW with a learning rate of $5 \times 10^{-5}$, a weight decay of $1 \times 10^{-4}$, and a batch size of 64. To prevent overfitting, early stopping is triggered after a patience of 20 epochs, with a maximum training duration of 200 epochs. The drug and protein encoders generate embeddings of size 768 and 1280, respectively, which are subsequently projected into a shared 1024-dimensional latent space via linear transformation. For interaction modeling, a bilinear attention module with 8 heads is employed, alongside a global dropout rate of 0.05.

\begin{table}[!t]
\caption{Experimental dataset statistics}
\label{tab:datasets}
\begin{tabular}{@{}llllllrrrrr@{}}
\toprule
\textbf{Dataset} & \textbf{\# Drugs} & \textbf{\# Proteins} & \textbf{\# Interactions} \\
\midrule
\multicolumn{4}{@{}l}{\textbf{Drug-Target Interaction}} \\
BindingDB\cite{Liu2025} & 14{,}643 & 2{,}623 & 49{,}199 \\
BioSNAP\cite{zitnik2018biosnap}       &  4{,}510 & 2{,}181 & 27{,}464 \\
Human\cite{liu2015credible,chen2020transformercpi}           &  2{,}726 & 2{,}001 &  6{,}728 \\
\midrule
\multicolumn{4}{@{}l}{\textbf{Drug Virtual Screening}} \\
LIT-PCBA\cite{tran2020litpcba}        & 415{,}225 & 15 & -- \\
\bottomrule
\end{tabular}
\begin{tablenotes}[flushleft]\footnotesize
  \item \textbf{Notes.} BioSNAP and Human sets originate from Huang~et~al.~(2021)~\cite{huang2021moltrans} and Liu~et~al.~(2015)~\cite{liu2015credible}, respectively. The BindingDB subset is a low-bias version compiled from the original database:  (\textit{i})~only pairs with IC$_{50}<100$\,nM (positive) or IC$_{50}>10\,000$\,nM (negative) are kept, yielding a 100-fold margin to suppress label noise;  (\textit{ii})~drugs exhibiting exclusively positive or negative pairs are removed to mitigate ligand-level bias.
\end{tablenotes}
\end{table}

\begin{table*}[t]
\centering
\caption{Performance comparison of DTI prediction methods on BindingDB, Human, BioSNAP datasets}

\label{tab:dti}
\resizebox{\linewidth}{!}{%
\begin{tabular}{lccccccccc}
\toprule
\multirow{2}{*}{\textbf{Method}} & \multicolumn{3}{c}{\textbf{BindingDB}} & \multicolumn{2}{c}{\textbf{Human}} & \multicolumn{3}{c}{\textbf{BioSNAP}} \\
\cmidrule(lr){2-4}\cmidrule(lr){5-6}\cmidrule(lr){7-9}
& \textbf{AUROC ↑ }& \textbf{AUPRC ↑ }&\textbf{ Accuracy ↑ }&\textbf{ AUROC ↑ }&\textbf{ AUPRC ↑ }&\textbf{ AUROC ↑ }& \textbf{AUPRC ↑ }&\textbf{ Accuracy ↑}\\
\midrule
SVM\cite{cortes1995support} & .939 ± .001 & .928 ± .002 & .825 ± .004 & .940 ± .006 & .920 ± .009 & .862 ± .007 & .864 ± .004 & .777 ± .011  \\
RF\cite{ho1995random} & .942 ± .011 & .921 ± .016 & .880 ± .012 & .952 ± .011 & .953 ± .010 & .860 ± .005 & .886 ± .005 & .804 ± .005   \\
DeepConv-DTI\cite{lee2019deepconv} & .945 ± .002 & .925 ± .005 & .882 ± .007 & .980 ± .002 & .981 ± .002 & .886 ± .006 & .890 ± .006 & .805 ± .009 \\
GraphDTA\cite{nguyen2021graphdta} & .951 ± .002 & .934 ± .002 & .888 ± .005 & .981 ± .001 & \underline{.982 ± .002} & .887 ± .008 & .890 ± .007 & .800 ± .007 \\
MolTrans\cite{huang2021moltrans} & .952 ± .002 & .936 ± .001 & .887 ± .006 & .980 ± .002 & .978 ± .003 & .895 ± .004 & .897 ± .005 & .825 ± .010  \\
DrugBAN\cite{bai2023interpretable} & \underline{.960 ± .001} & \textbf{.948 ± .002} & \textbf{.904 ± .004} & \underline{.982 ± .002} & .980 ± .003 & .903 ± .005 & .902 ± .004 & .834 ± .008  \\
SiamDTI\cite{zhang2024crossfield} & \textbf{.961 ± .002} & \underline{.945 ± .002} & .890 ± .006 & .970 ± .002 & .969 ± .003 & \underline{.912 ± .005} & \underline{.910 ± .003} & \underline{.855 ± .004} \\
\midrule
\textbf{SaBAN (Ours)} & .957 ± .002 & \textbf{.948 ± .001} & \underline{.902 ± .003} & \textbf{.983 ± .003} & \textbf{.984 ± .006} & \textbf{.930 ± .003} & \textbf{.929 ± .002} & \textbf{.858 ± .006} \\
\bottomrule
\end{tabular}%
}
\begin{tablenotes}[flushleft]\footnotesize
  \item \textbf{Notes.} Best values are in \textbf{bold}; Second-best are \underline{underlined}; Evaluation metrics include Area Under the Receiver Operating Characteristic Curve (AUROC), Area Under the Precision-Recall Curve (AUPRC), and accuracy. Model selection was based on the highest validation AUROC, and final results were averaged over five-fold cross-validation.
\end{tablenotes}
\end{table*}

\begin{table}[t]
\centering
\caption{Performance comparison on LIT-PCBA benchmark}

\resizebox{\columnwidth}{!}{
\begin{tabular}{lccccc}
\toprule
\multirow{2}{*}{\textbf{Method}} &
\multirow{2}{*}{\textbf{AUROC (\%) ↑}} &
\multirow{2}{*}{\textbf{BEDROC (\%) ↑}} &
\multicolumn{3}{c}{\textbf{EF ↑}} \\
\cmidrule(lr){4-6}
 & & & \textbf{0.5\%} & \textbf{1\%} & \textbf{5\%} \\
\midrule
Surflex\cite{spitzer2012surflex} & 51.47 & - & - & 2.50 & - \\
Glide-SP\cite{halgren2004glide} & 53.15 & 4.00 & 3.17 & 3.41 & 2.01 \\
Planet\cite{zhang2023planet} & 57.31 & - & 4.64 & 3.87 & \underline{2.43} \\
GNINA\cite{mcnutt2021gnina} & \underline{60.93} & 5.40 & - & 4.63  & - \\
DeepDTA\cite{ozturk2018deepdta} & 56.27 & 2.53 & - & 1.47  & - \\
BigBind\cite{brocidiacono2022bigbind} & 60.80 & - & - & 3.82 & - \\
DrugCLIP\cite{drugclip} & 57.17 & \underline{6.23} & \underline{8.56} & \underline{5.51} & 2.27 \\
\midrule
\textbf{SaBAN (Ours)}  & \textbf{68.16} & \textbf{13.35} & \textbf{9.17} & \textbf{6.33} & \textbf{3.36} \\
\bottomrule
\end{tabular}
}
\label{tab:pcba}
    \begin{tablenotes}[flushleft]\footnotesize
      \item \textbf{Notes.} Best values are in \textbf{bold}; second-best are \underline{underlined}. For virtual screening, the Boltzmann-Enhanced Discrimination of Receiver Operating Characteristic (BEDROC) and Enrichment Factor (EF) were reported to capture early retrieval performance. Model selection was based on the highest validation AUROC, and final results were averaged over five-fold cross-validation.
    \end{tablenotes}
\end{table}

\section{Results \& Discussion}
\label{sec:res}

\subsection{Performance on DTI}
Across three DTI datasets, the model achieves state-of-the-art results on Human and BioSNAP benchmarks. On BindingDB, performance aligns closely with DrugBAN and remains competitive with leading baselines. These results demonstrate consistent generalization and validate the effectiveness of the proposed architecture, as shown in Table~\ref{tab:dti}.

\subsection{Performance on Virtual Screening}
On the LIT-PCBA benchmark for virtual screening, the proposed method outperforms the previous state-of-the-art, DrugCLIP, as shown in Table~\ref{tab:pcba}. SaBAN achieves an AUROC of 68.16\%, yielding a 10.99\% absolute improvement. For early recognition, SaBAN attains a BEDROC of 13.35\%, exceeding DrugCLIP by more than twofold.

\subsection{Ablation Study}
\begin{figure}[t]
    \centering
  \includegraphics[width=\linewidth]{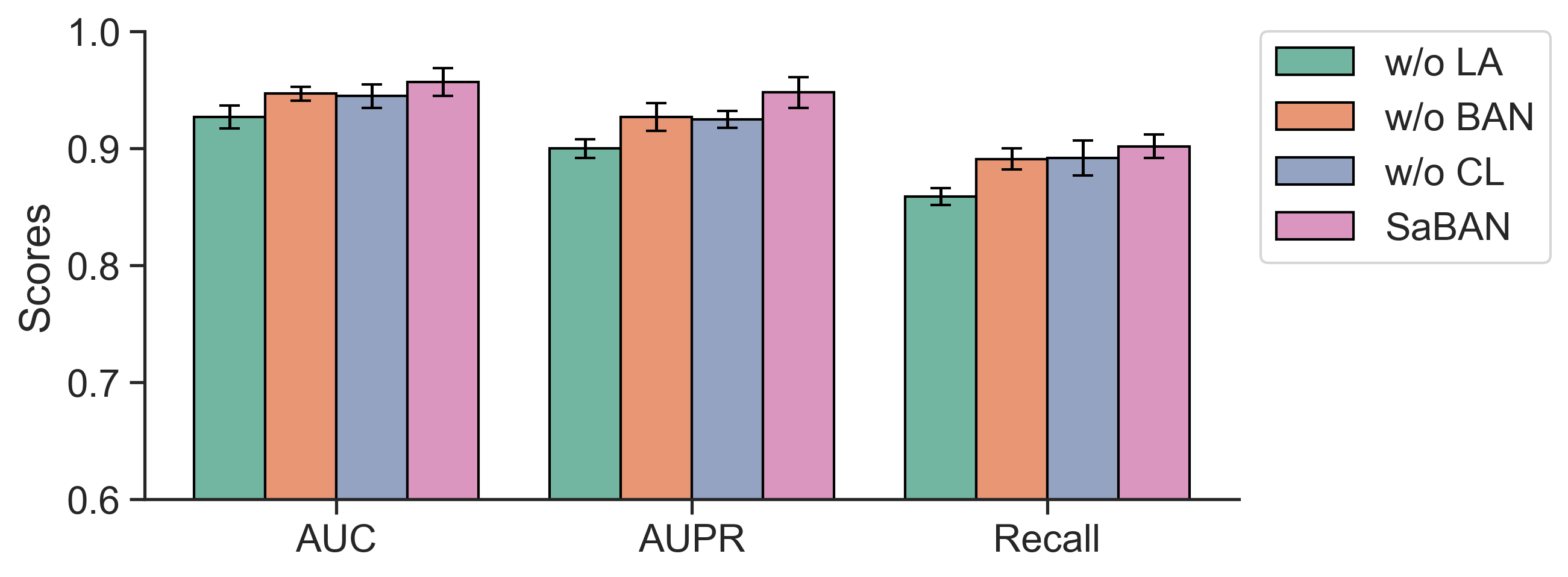}
  \caption{Comparison between SaBAN and its variants on BindingDB dataset.}
  \label{fig:ablation}
\end{figure}

To quantify the contribution of each module, we conducted ablation experiments by removing the learned aggregation layer (LA), the bilinear attention network (BAN), and the contrastive learning (CL) components. As shown in Fig.~\ref{fig:ablation}, removing LA results in the largest performance degradation across all metrics. Excluding BAN or CL also leads to noticeable declines, confirming that both interaction modeling and representation regularization are important for robustness.
\subsection{Visualization}
\begin{figure}[htpb]
    \centering
  \includegraphics[width=\linewidth]{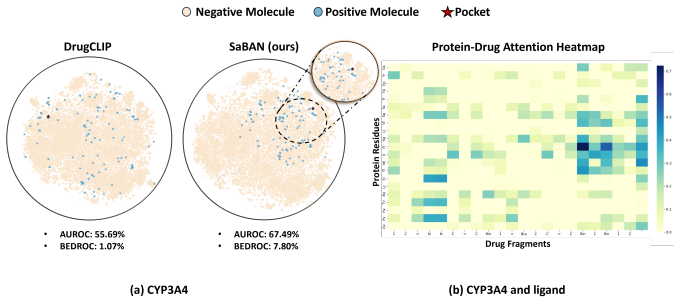}
  \caption{Visualization of feature distributions for the target CYP3A4 (a) using DrugCLIP and SaBAN models and the attention map (b) of its ligand.}
  \label{fig:t-sne}
\end{figure}

We visualize the embedding distribution for the CYP3A4 target and its attention profile to interpret the representations. Fig.~\ref{fig:t-sne}(a) shows the T-SNE projection, where SaBAN produces embeddings with positive samples positioned closer to the binding pocket than DrugCLIP, indicating improved structural correspondence. Fig.~\ref{fig:t-sne}(b) displays the ligand-level attention weights, concentrating on functional residues within the pocket and providing evidence of model interpretability.
\section{Conclusion}
\label{sec:conclusion}
We propose a contrastive framework that combined cross-modal alignment with bilinear interaction modeling to extract drug–pocket relationships from tokenized sequences. The approach delivered state-of-the-art results on Human and BioSNAP datasets and maintained competitive performance on BindingDB. In virtual screening evaluations, it outperformed previous methods on LIT-PCBA, achieving notable improvements in AUROC and BEDROC. These capabilities supported efficient pre-screening using accessible one-dimensional sequences augmented with novel structure-aware representations.

\section{ACKNOWLEDGEMENTS}
This work was supported by an internal grant from The Hong Kong Polytechnic University (Project No. P0051278, Jung Sun Yoo) and the General Research Fund (Project No. PolyU 15101422, Jung Sun Yoo) from the Research Grants Council of the Hong Kong Special Administrative Region, China.

\AtBeginEnvironment{thebibliography}{%
    \fontsize{10}{11}\selectfont  
    \setlength{\itemsep}{0pt}
    \setlength{\parskip}{0pt}
}
\bibliographystyle{IEEEbib}
\bibliography{refs}

\end{document}